\documentclass[a4paper, 10pt, conference]{ieeeconf}      
\usepackage{amsfonts}
\usepackage{amsmath}
\usepackage{graphicx}
\usepackage{subfig}
\usepackage{xcolor}
\usepackage{pifont}
\usepackage{fancyhdr}
\usepackage{graphicx}
\usepackage{amsmath}
\usepackage{cite}
\DeclareUnicodeCharacter{2009}{\,} 
\IEEEoverridecommandlockouts                              
\title{\LARGE \bf
Sheep Facial Pain Assessment Under Weighted Graph Neural Networks 
}

\author{\parbox{16cm}{\centering
    {\large Alam Noor$^{\Phi,\Psi}$, Luis Almeida$^\Psi$, Mohamed Daoudi$^{\ddag,\ast}$, Kai Li$^\Phi$ and Eduardo Tovar$^\Phi$}\\
    {\normalsize
    $^\Phi$ CISTER Research Center, Porto, Portugal\\
    $^\Psi$ Faculty of Engineering University of Porto, Portugal\\
    $^\ddag$ Univ. Lille, CNRS, Centrale Lille, UMR 9189 CRIStAL, F-59000, Lille, France \\
    $^\ast$ IMT Nord Europe, Institut Mines-Télécom, Univ. Lille, Centre for Digital Systems, F-59000 , Lille, France}\\
    }
}

\begin{document}

\pagestyle{plain}
\maketitle 
\thispagestyle{fancy}

\begin{abstract}
Accurately recognizing and assessing pain in sheep is key to discern animal health and mitigating harmful situations. However, such accuracy is limited by the ability to manage automatic monitoring of pain in those animals. Facial expression scoring is a widely used and useful method to evaluate pain in both humans and other living beings. Researchers also analyzed the facial expressions of sheep to assess their health state and concluded that facial landmark detection and pain level prediction are essential. For this purpose, we propose a novel weighted graph neural network (WGNN) model to link sheep's detected facial landmarks and define pain levels. Furthermore, we propose a new sheep facial landmarks dataset that adheres to the parameters of the Sheep Facial Expression Scale (SPFES). Currently, there is no comprehensive performance benchmark that specifically evaluates the use of graph neural networks (GNNs) on sheep facial landmark data to detect and measure pain levels. The YOLOv8n detector architecture achieves a mean average precision (mAP) of 59.30\% with the sheep facial landmarks dataset, among seven other detection models. The WGNN framework has an accuracy of 92.71\% for tracking multiple facial parts expressions with the YOLOv8n lightweight on-board device deployment-capable model. 
\end{abstract}

\section{INTRODUCTION}
In the agriculture domain, raising herds plays a significant role. In particular, sheep herds are very common in many cultures and have an important economic and social impact \cite{SMITH2024106194}. The success of raising herds depends on the well-being of the animals; thus, tracking the animals condition, sheep in particular, becomes very relevant. One specific metric that has been used is the animal pain level, estimated from images of the animal face resorting to  mobile devices, such as Unmanned Aerial Vehicles (UAVs) equipped with on-board cameras \cite{10155900}. Distinct facial parts and their interactions (facial expressions) communicate different emotions and facilitate the detection of possible health problems in the natural state \cite{NOOR2023100366, NOOR2020105528,10388128}. This information can be used to develop early diagnosis strategies and avoid the spread of sickness to other sheep or animals in the farm.

Recent attempts at facial and pose recognition and identification and classification of sheep health condition rely on deep learning \cite{GUO2023108027, GU2023108143, SARWAR2021106219, LI2023107651, HITELMAN2022106713, ZHANG2022107452, 10603691}. All of these studies focus on whole-face detection and pose estimation, rather than assessing individual facial landmarks to determine facial expressions. Different parts of the sheep's face, like the ears, eyes and nose, allow establishing a relationship between pain levels and certain expressions. However, analyzing the facial expression from the whole face image leads to less accurate models for the  facial pain and pose assessment. Moreover, none of these models can accurately detect the facial landmarks of sheep, which is crucial for a comprehensive assessment of the animals' health conditions.

Therefore, we face two challenges: first analyzing the facial expressions of sheep to develop a dataset of their facial landmarks, and second defining the overall pain level of the sheep's face by analyzing each part of the face and avoiding landmarks that do not exhibit any abnormalities. Moreover, it is important to identify the sheep facial landmarks using the predefined parameters of the Sheep Pain Facial Expression Scale (SPFES), which evaluates each facial expression landmark (ears, eyes, and nose) in its natural environment \cite{MCLENNAN201619, NOOR2023100366, 7961768} and using mobile devices like, UAVs \cite{SARWAR2021106219}. 

\begin{figure*}[htbp!]
    \begin{center}
    \hspace*{0em}
        \includegraphics[scale=0.30]{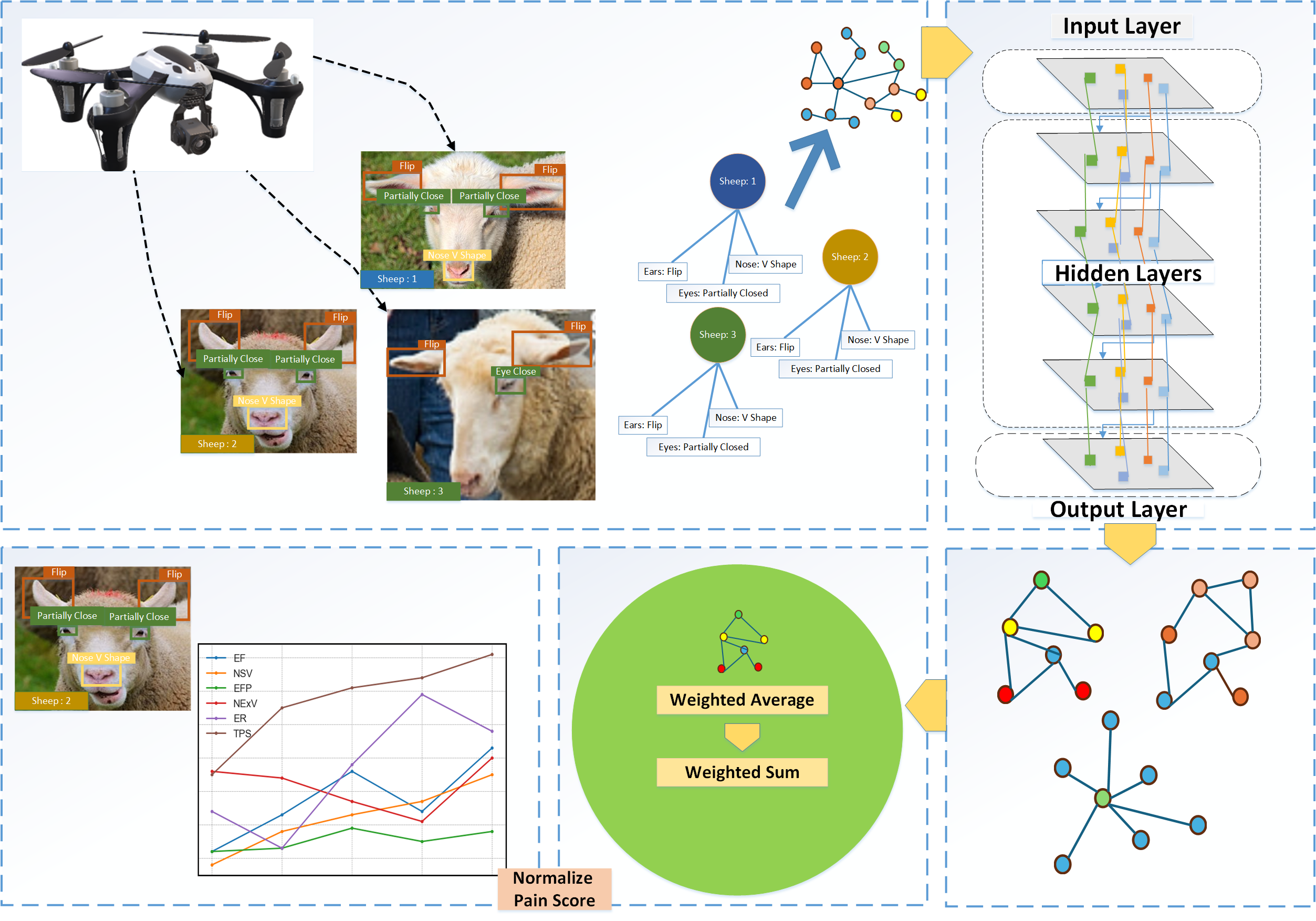}
        \caption{The proposed WGNN model workflow for identifying sheep facial landmarks involves mobile devices identifying the sheep's facial features, which then undergo evaluation by a lightweight object detector model. Following that, the WGNN calculates the overall pain score by clustering the various landmarks on the sheep's face.}
        \label{SheepGNN}
    \end{center}
\end{figure*}

To address these issues, we propose a Weighted Graph Neural Network (WGNN) model for health condition decision that receives input from an object detector lightweight YOLO model based on five facial parts using SPFES as shown in Fig. \ref{SheepGNN}. We are motivated by the progress of deep neural networks that led to an extensive use of Graph Neural Networks (GNNs) in complicated connection link prediction. Firstly, a GNN-based prediction method learns the similarity of facial landmarks as node representations for link prediction, using detector prediction in the background. Secondly, a GNN updates and learns node representations by exchanging SPFES information for each part of the face about nearby facial landmarks on a regular basis to provide accurate health condition output.  

To validate the proposed study, we developed a sheep facial landmarks dataset using SPFES parameters. The facial image parameters are taken from the dataset used in a previous work \cite{NOOR2020105528} to validate the proposed study. Moreover, in this paper, we propose a WGNN model that defines the pain level of the sheep face expression by finding the relationship among the facial parts and clustering the related ones. A weighted average aggregated score that is based on SPFES roles processes each cluster of the GNN model to assign weight for pain assessment. 

In summary, the main contributions of our work are the following: 

\begin{enumerate}
    \item We propose a novel WGNN model that takes the input detection algorithm eyes, ears, and nose for facial expression and creates clusters of similar or closer pain scores of the facial parts expression. The WGNN model utilizes a probability distribution to determine the optimal graph structure, which allows it to observe the closest and most similar nodes and edges, thereby producing accurate facial expressions. Moreover, the proposed WGNN model develops a weighted average of the pain levels of the parts expression within that cluster to define the total pain in normalized form. 
    \item We propose a new sheep facial landmark dataset to support the proposed WGNN model in which every part of the sheep face has a bounding box with nine different facial expressions using five facial parts: eye constraint, cheeks stretching, ears positioning, lips and jaws shape, and nose posture according to the standard of SPFES parameters. 
    \item We conduct experiments to assess the effectiveness of the proposed WGNN model in real-world scenarios, utilizing eight different state-of-the-art detection algorithms to identify different facial expressions in sheep. Four top mAP score detection algorithms utilize the proposed WGNN model. The experimental results show that the proposed WGNN model achieves the highest accuracy (92.71\%) with YOLOv8 (having mAP@59.30\%) than state-of-the-art models in comparison.
\end{enumerate}

The paper is organized as follows: Section \ref{relatedwork} reviews related work, highlighting and discussing the deep learning model used on sheep facial expression and sheep face datasets, as well as sheep body detection and preprocessing methods. Section \ref{systemmodel} evaluates the proposed WGNN model and explains its entire architecture. Section \ref{ExperimentalWork} provides a discussion of the experimental setup, findings, and limitations. Finally, Section \ref{Conclusion} concludes the paper and outlines the advantages of the proposed WGNN model.

\section{Related Work}\label{relatedwork}
In the past, researchers have studied the recognition and expression of animal faces to identify their behaviors and illnesses. These animals, particularly sheep, exhibit behavior more akin to that of humans \cite{NOOR2023100366}. These behaviors are the result of sheep's natural ability to evade potential threats and problems, as well as their human similarity, such as group living and the ability to recognize their mother and siblings, even after a long period of separation. These two factors have a significant impact on sheep's behavior and responses in a variety of situations. 

The literature studies several deep learning approaches for sheep face classification and detection. In particular, Hao et al. \cite{agriculture14030468} studied the sheep facial expression using the Single Shot MultiBox Detector (SSD) algorithm to detect the whole face of sheep and define its expressions. Ying et al. \cite{GUO2023108027} presented the detection of sheep breed using DT-YOLOv5 and were capable of recognizing facial features. Zishuo et al. \cite{GU2023108143} introduced a procedure that involves two phases: identification and classification. A detection network determines if each sheep's activity is normal, physiological, or disruptive, while traditional networks multi-scale feature aggregation, attention mechanism, and depthwise convolution module are used to let the network balance model size and detection accuracy. Farah et al. \cite{SARWAR2021106219} presented single as well as seven-layer convolutional neural network (CNN) models with the help of centroids by UAVs to identify the sheep. Furthermore, after fine-tuning the pre-trained models, they assembled the FCN and defined models to address recall and precision issues, while Li et al. \cite{LI2023107651} combined CNNs and vision transformers to recognise the sheep faces by extracting feature representations. Hitelman et~al. \cite{HITELMAN2022106713} studied face detection and classification using a biometric identification model. They applied faster R-CNN to detect the sheep's face in an image, while pretrained models were used to identify the face into seven distinct classification models. Zhang et~al. \cite{ZHANG2022107452} used the YOLOv4 model to identify sheep and added the convolutional block attention module (CBAM) to make the extraction of model features more stable with the help of the biometric system. In another work, Zhang et~al. \cite{10.1093/jas/skae066} used YOLOv7 with multiple attention mechanisms to detect the sheep faces. Additionally, the same authors \cite{10342186} used the YOLOv7-tiny for the multiple sheep abnormal behaviors using sheep whole body structure from a video camera. 

Similarly to the previous works, Bati et~al. \cite{Bati2024} studied the YOLOv5 model with SORT algorithm to detect the whole body of the sheep to identify and track the animal behaviors. Zhang et~al. \cite{ani13111824} also introduced YOLOv5 for the detection of the whole face of the sheep, while Ayub et~al. \cite{10469582} used YOLOv5s to classify the corresponding activity state in active and non-active for the sheep body. Zhang et~al. \cite{ZHANG2024108697} also introduced the data called multi-view sheep face images and applied the vision transform model to recognize the sheep faces with a heat map. Kelly et~al. \cite{KELLY2024110027} presented a sheep activity dataset to analyze the behaviors of the sheep with deep learning pre-trained models. Xue et~al. \cite{ani14131923} introduced a sheep face orientation recognition algorithm to different face orientations and a feature point-matching and reconstructing the sheep face. However, the algorithm requires images from different angles to accurately recreate the sheep face. Pang et~al. \cite{Pang2023} studied the feature extraction of sheep faces using an attention residual module to aggregate the features received from the input of a spherical camera to capture image data. However, the model relies on classification, making it challenging to identify sheep faces for its application. Cai et~al. \cite{10.1145/3650400.3650652} used convolutional neural networks for the sheep face recognition for disease prevention, while Xinyu et~al. \cite{10505584} presented the dilated convolutional attention module for gender identification using sheep faces as a binary classification model.

The existing models, as illustrated in the literature, focus primarily on the sheep's entire face or body. With such an approach, classifying the sheep's facial pain is very difficult. To classify the pain, we must examine every part of the sheep's entire face. Furthermore, the detection models only detect facial parts expression, requiring the clustering of painful parts and their separation from non-painful ones to define and combine the total face pain. Therefore, our study follows the novel sheep facial landmark dataset, proposing the use of WGNN to determine the total pain of the sheep faces.

\section{System Model}\label{systemmodel}
Nodes of the GNN model serve as representations of the eyes, ears, and nose for facial expression understanding, and their relations are defined as edges. Given a complete SPFES graph that includes all possible relationships among the eyes, ears, and nose, we want to automatically infer a parse graph by keeping the meaningful edges and labeling the nodes. We consider $\mathcal {G} = (\mathcal {N}, \mathcal {E}, \mathcal {O})$ denotes the complete SPFES graph. Node $n \in \mathcal {N}$ takes unique values from $\{1, \cdots, |\mathcal {N}|\}$. Edges $e \in \mathcal {E}$ are two-tuples $e = (n, w) \in \mathcal {N} \times \mathcal {N}$. Each node $n$ has an output state $o_n \in \mathcal {O}$ that takes a value from a set of labels
$\{1, \cdots, \mathcal {O}_n\}$ (e.g., facial actions). These relationships are between the facial parts expressions, and they are grouped based on the presence and severity of pain. The GNN outputs $\mathcal {O}$ state clusters, $(\mathcal {C}_1, \mathcal {C}_2, \ldots, \mathcal {C}_o)$ with each cluster \(\mathcal {C}_j\) containing a subset of the 9 facial expressions. For every part of the facial expressions, YOLO detects the pain levels ($p_i \in \{0, 1, 2\}$) for all the 9 facial parts expressions $\mathcal {P} = [p_1, p_2, p_3, \ldots, p_9]$.

Therefore, in the graph $\mathcal {G}$, $\mathcal {N}$ represents the 9 facial part expressions and $\mathcal {E}$ presents relationships or similarities between facial part expressions $\mathcal {P}$. We can determine these based on their proximity to the face structure and similar pain scores. Face structure proximity refers to how physically close facial parts are in the sheep's anatomy. Moreover, physically closer facial parts, such as cheeks to the nose and lips to the jaw, are more likely to have a direct relationship or influence each other. The nodes in $\mathcal {N}$ that are close in the face structure will have an edge in $\mathcal {E}$ connecting them, indicating a direct relationship or interaction. Then $\mathcal {P}$ should indicate comparable levels of $p_i$ in the facial parts. Facial areas with similar $p_i$ ratings might show that they suffer from similar degrees of stress or discomfort, potentially due to a prevalent underlying cause or condition. Thus, within the graph, nodes in $\mathcal {N}$ with corresponding $p_i$ ratings will be linked by $\mathcal {E}$ signifying a likeness in their condition to create a cluster $\{1, \cdots, \mathcal {O}_n\}$ (e.g., facial actions).  

$\mathcal {N}$ has an associated feature vector $\psi$ representing various facial pain aspects of the animal condition. The initial feature vector for a facial part expression \(i\) is denoted as \(\psi_i^{(0)}\). For facial part expressions type ($FPT$), the $\psi$ can be represented as in \eqref{psi}:
\begin{equation}\label{psi}
    \psi_i^{(0)} = [p_i, \text{FPT}_i]
\end{equation}

Next, for each, $FPT$ a parse graph $g = (\mathcal {N}_g, \mathcal {E}_g, \mathcal {O}_g)$ is constructed, which is a sub-graph of $\mathcal {G}$,
where, $\mathcal {N}_g \subseteq \mathcal {N}$ and $\mathcal {E}_g \subseteq \mathcal {E}$ \cite{10.1007/978-3-030-01240-3_25}. Given the node features $\psi^\mathcal {N}$ and edge features $\psi^\mathcal {E}$, we
want to infer the optimal parse graph $g^*$ that best explains the facial expression according
to a probability distribution $P_d$:

\begin{equation}\label{g}
\begin{split}
    g^* = \arg\max_g P_d(\mathcal {O}_g \mid \mathcal {N}_g, E_g, \psi) P_d(\mathcal {N}_g, E_g \mid \psi, \mathcal {G}),
\end{split}
\end{equation}
where $\psi = \{\psi^\mathcal {N}, \psi^E\}$, $P_d(\mathcal {N}_g, E_g \mid \psi, \mathcal {G})$ evaluates the graph structure and
$P_d(\mathcal {O}_g \mid \mathcal {N}_g, E_g, \psi)$ is the labeling probability for the nodes in the parse graph. Once the (\ref{g}) define the optimal graph structure to observe the best $\mathcal {O}_g$ having proximity and similarity nodes and edges, WGNNs update each $\psi^\mathcal {N}$ by aggregating information from its neighbors with $\psi^\mathcal {E}$ through multiple layers with using aggregation and update strategy with each node aggregating messages from its neighbors. The message from a neighboring node \(j\) to node \(i\) in the layer \(k\) is \(\psi_j^{(k-1)}\) to make a precise cluster for the degree of pain level, and the aggregate message \(\mathcal{M}_i^{(k)}\) for the node \(i\) is given in \eqref{M},
\begin{equation}\label{M}
    \mathcal{M}_i^{(k)} = \sum_{j \in \mathcal{N}(i)} \psi_j^{(k-1)},
\end{equation}
where \(\mathcal{N}(i)\) denotes the set of neighbors of node \(i\). The node feature \(\psi_i\) is then updated using the aggregated message by \eqref{psi_k},
\begin{equation}\label{psi_k}
    \psi_i^{(k)} = \mathcal {F}(\sigma) (\mathcal{W}^{(k)} \mathcal{M}_i^{(k)} + \mathbf{b}^{(k)}),
\end{equation}
where $\mathcal {F}(\sigma)$ is an ReLU activation function, \(\mathcal{W}^{(k)}\) is the weight matrix for the \(k\)-th layer, and \(\mathbf{b}^{(k)}\) is the bias vector for the \(k\)-th layer. After $\mathcal{K}$ layers of message passing, final node embedding \(\psi_i^{(k)}\) used for to embedding to cluster the nodes (facial parts expressions) into $\mathcal {O}$ clusters, and for each cluster \(\mathcal {C}_j\), calculate an aggregated pain score using weighted average of the pain levels of the parts expressions within that cluster as mentioned in \eqref{S_j},

\begin{equation}\label{S_j}
    \mathcal {S}_j = \frac{\sum_{i \in C_j} w_i \cdot p_i}{\sum_{i \in C_j} w_i},
\end{equation}
while, $w_i$ is the weight assigned to each part expressions $p_i$ based on the SPFES. After all, we sum up the aggregated pain scores $\mathcal {S}_j$ using \eqref{T_p} to compute the final pain score for the entire face with weight $w_j$ to each cluster based on its relative importance in the SPFES,
\begin{equation}\label{T_p}
    \mathcal {T}_p = \sum_{j=1}^o w_j \cdot \mathcal {S}_j
\end{equation}
In last, the proposed model uses a normalization function $NPS(.)$ with maximum possible pain $\mathcal {T}_{max}$ for the whole sheep face to ensure that the facial pain falls in the range from 0 to 100\% as given in \eqref{NPS}. 
\begin{equation}\label{NPS}
    \mathcal{NPS} = \left( \frac{\mathcal {T}_p}{\mathcal {T}_{max}} \right) \times 100
\end{equation}

The cross-entropy used in $\mathcal {O}$ clusters classification is define as follows, where \( \mathcal {U}_p \in \Sigma^{\langle o \rangle} \) shows the actual probability distribution and \( \mathcal {V}_q \in \Sigma^{\langle o \rangle} \) shows the predicted probability distribution as given in the \eqref{CrossEntropyLoss},

\begin{equation}\label{CrossEntropyLoss}
\text{CrossEntropyLoss}(\mathcal {U}_p, \mathcal {V}_q) = - \sum_{i=1}^{o} \mathcal {U}_p \log \mathcal {V}_q
\end{equation}

The proposed WGNN model groups the painful facial features into clusters based on their levels of pain and adds up the pain scores for each cluster, allowing it to determine the overall degree of pain for the entire face. The formulation of the proposed model considers the relationships between facial part expressions, resulting in a more accurate assessment of the overall pain level.

\section{Experimental Work}\label{ExperimentalWork}
We conducted experiments on the sheep facial landmarks dataset that we present next, developed using SPFES parameters, to assess sheep facial pain. The dataset is divided into 80\% images with annotations for training and 20\% images for testing and validation. We meticulously label each image of the nose, ears, and eyes with boundary boxes to detect the level of pain. We trained the model using Pytorch on a GPU-equipped workstation with a GeForce RTX A3000. We chose different YOLO and other state-of-the-art models for training on the GeForce RTX A3000 with a batch size of 4. We utilize the Adam optimizer due to its swift and effortless convergence to the optimal point, which sets it apart from other optimizers for YOLO and other state-of-the-art models, and we employ cross-entropy loss to cluster the facial landmarks. The learning rate decay is set to 0.001, with $\beta$1 of 0.9 and $\beta$2 equal to 0.999. Every YOLO and other state-of-the-art model trained for 100 epochs with WGNN to achieve the desired accuracy.
\subsubsection{Dataset}
\begin{figure}[htbp!]
    \begin{center}
    \hspace*{-1em}
        \includegraphics[scale=0.11]{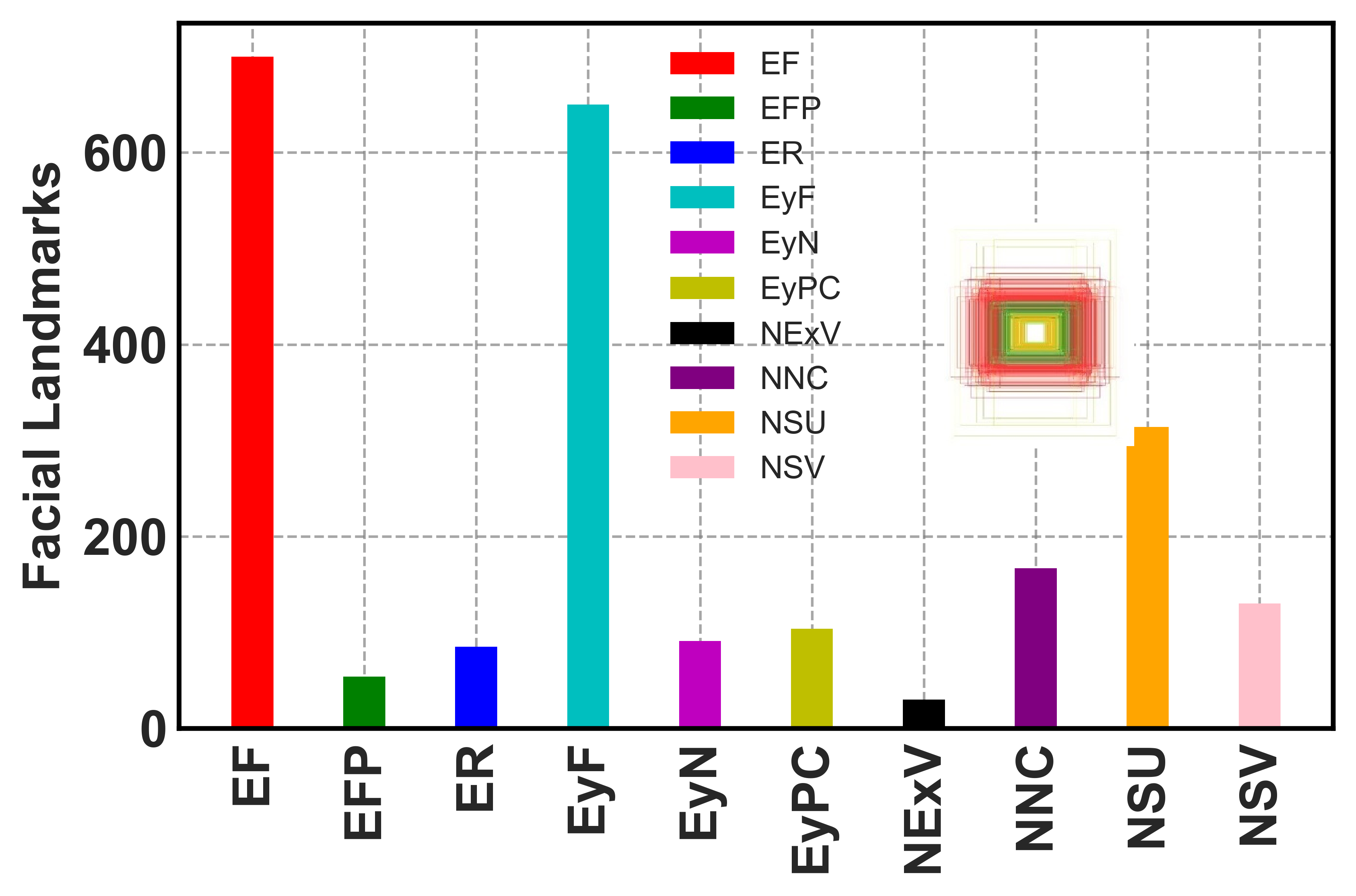}
        \caption{Sheep facial units statistical approach with bounding boxes for the evaluations in which flat ears (EF), ear flipped (EFP), ear rotation (ER), eyes fully opened (EyE), eyes not classified (EyN), eyes partially classified (EyPC), nose extended V shape (NExV), nose shallow U shape (NSU), nose not classified (NNC), and nose shallow V shape (NSV) are defined to classify different pain levels and their corresponding positions. The shape above the NNC represents sizes of different bounding boxes for facial landmarks. }
        \label{dataset}
    \end{center}
\end{figure}

We developed a dataset with a bounding box of sheep facial landmarks using high-resolution images from well-known sources according to the SPFES standard \cite{MCLENNAN201619, 7961768}. These sources are Mendeley \footnote{http://dx.doi.org/10.17632/y5sm4smnfr.5} and the study is carried out in \cite{NOOR2020105528}. We use the scale to evaluate expression in five facial parts: eye constraint, cheeks stretching, ears positioning, lips and jaws shape, and nose posture. We evaluate these parts expressions based on the presence or absence of abnormal expression, categorizing them as not present, slightly present, or substantial, as shown in Fig. \ref{dataset}.

We are looking into new ways of rethinking transformation to change the color background, such as contrast, flapping, equalizing, solarizing, and sharpness, so that we can add visual effects to the sheep images and reduce the overfitting \cite{9750351}. This is because deep learning architectures can handle changes in color, rotation, and axis-aligned bounding boxes. The rethinking transformation sets the random probability for all parameters to 0.5, except for solarization, which has a threshold of 128. The head of the sheep can move in a 45-degree clockwise and anti-clockwise rotation, which can vary depending on the rotation degree from the center (width and height by 2). We compute the new bounding box dimensions of the video frames after obtaining the sine and cosine from the rethinking transformation rotation matrix. 

\subsubsection{Results and Evaluation}
\begin{figure*}[htbp!]
\centering
\hspace*{0em}
\subfloat[]{\includegraphics[width=1.7in]{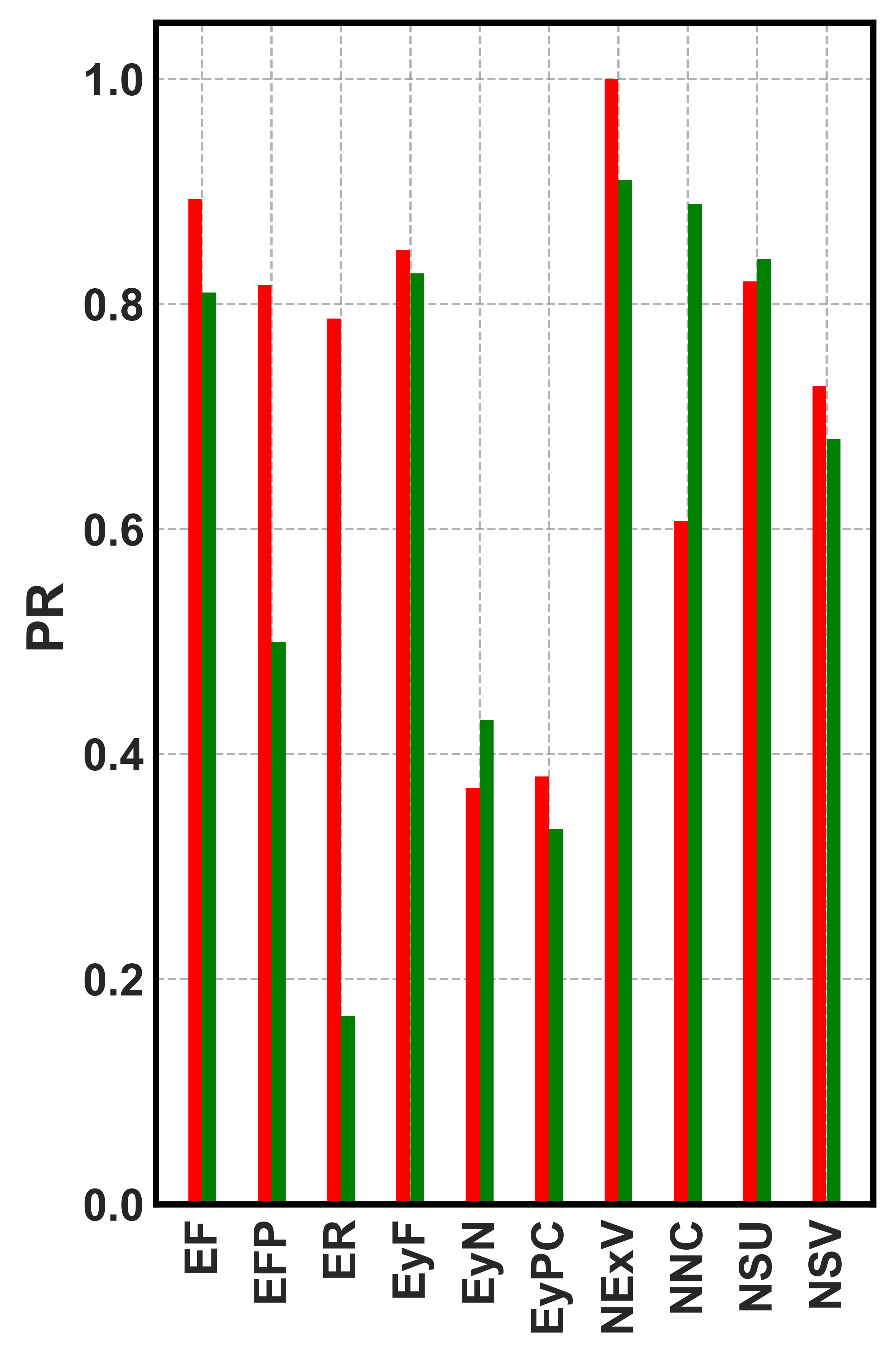}} 
\subfloat[]{\includegraphics[width=1.7in]{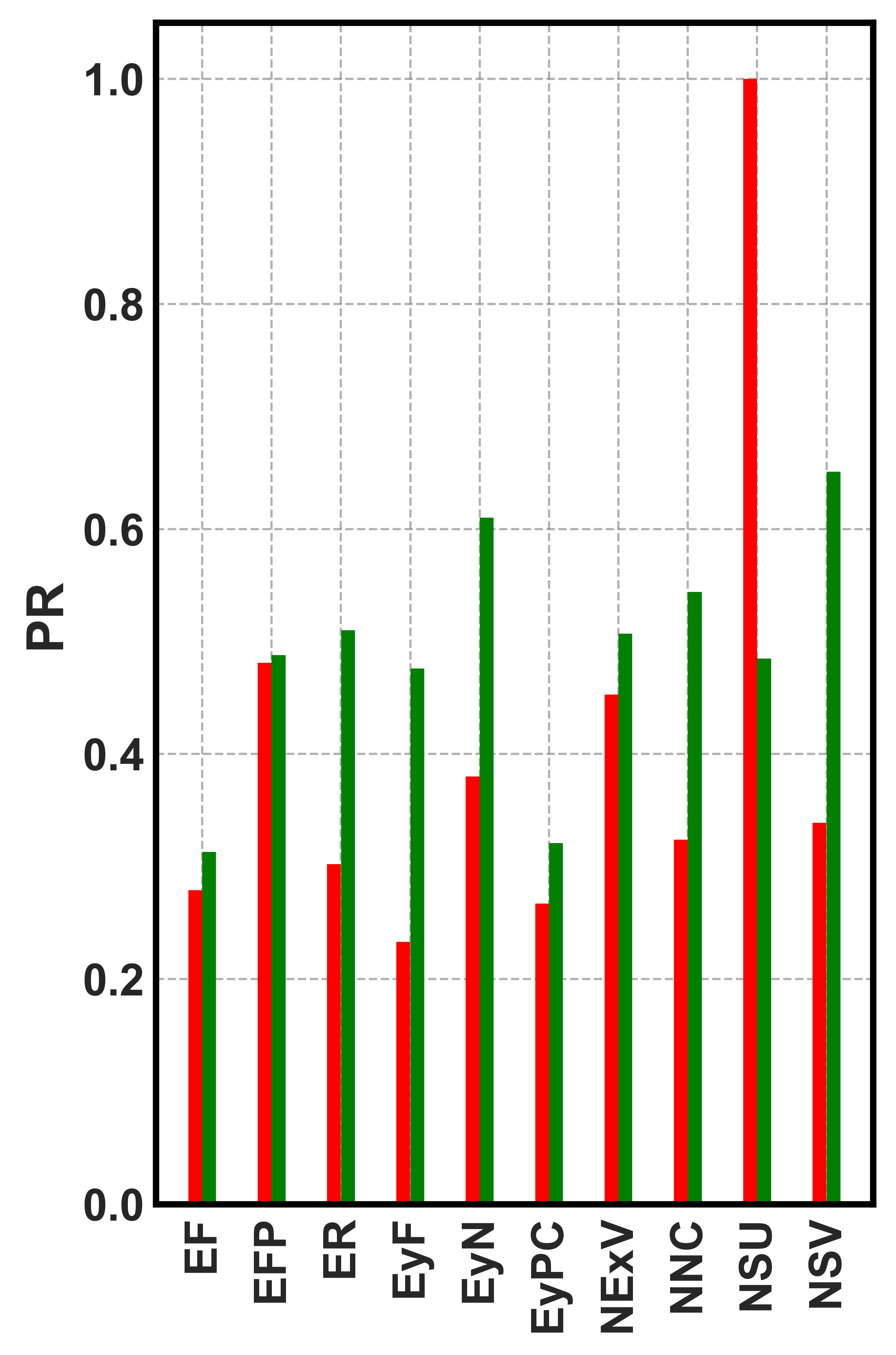}}
\subfloat[]{\includegraphics[width=1.7in]{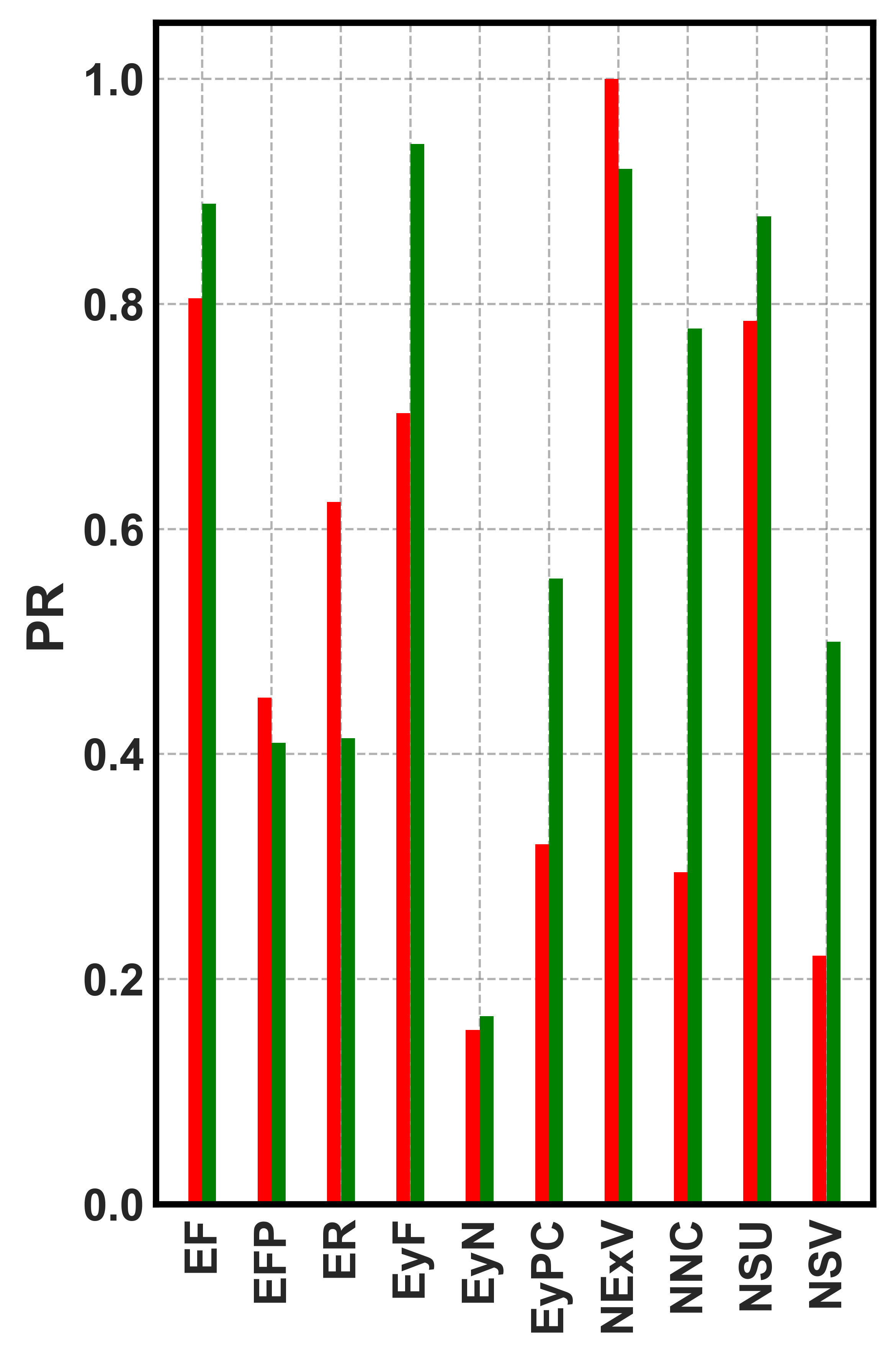}}
\subfloat[]{\includegraphics[width=1.7in]{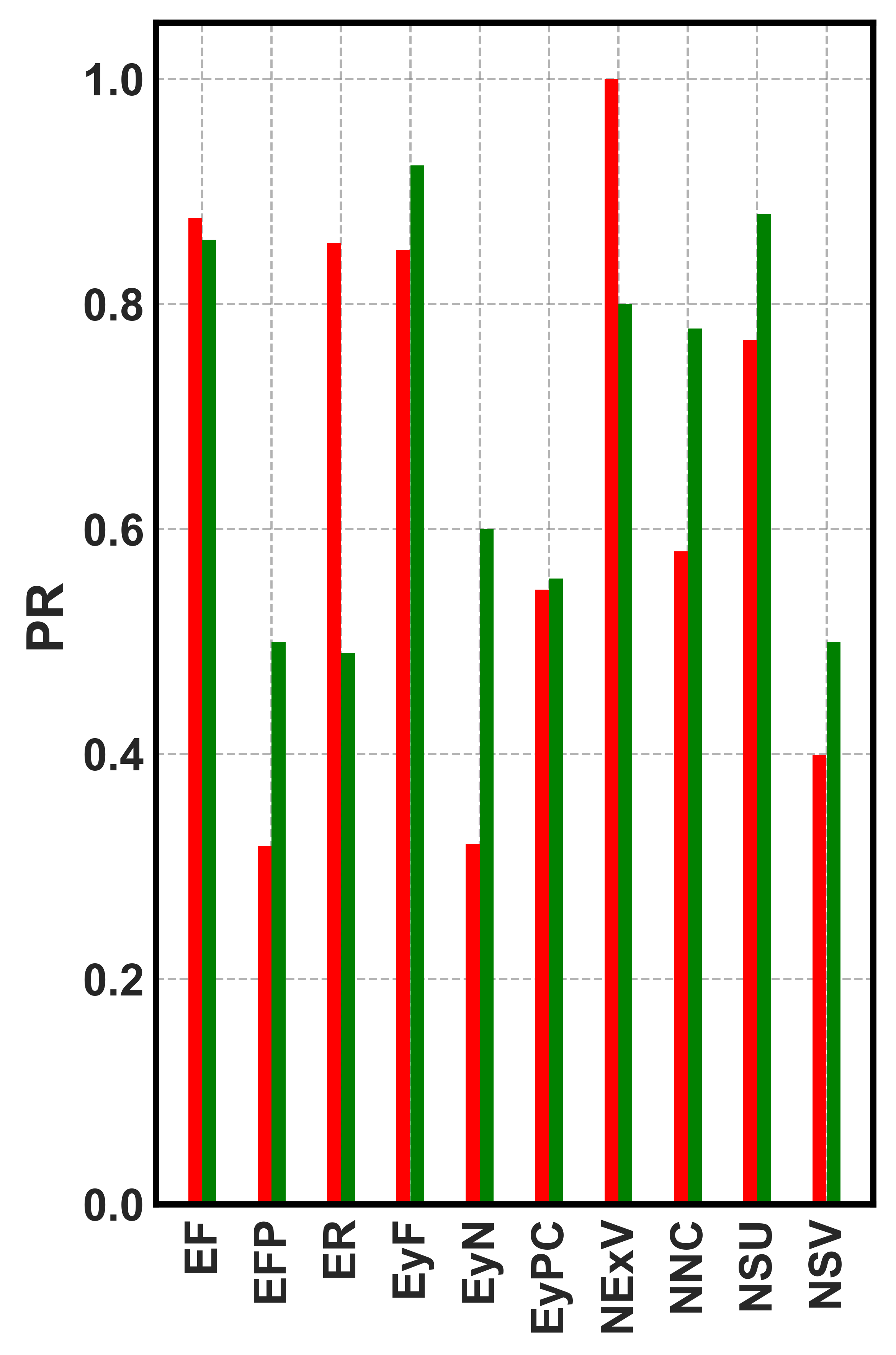}}\\
\subfloat[]{\includegraphics[width=1.7in]{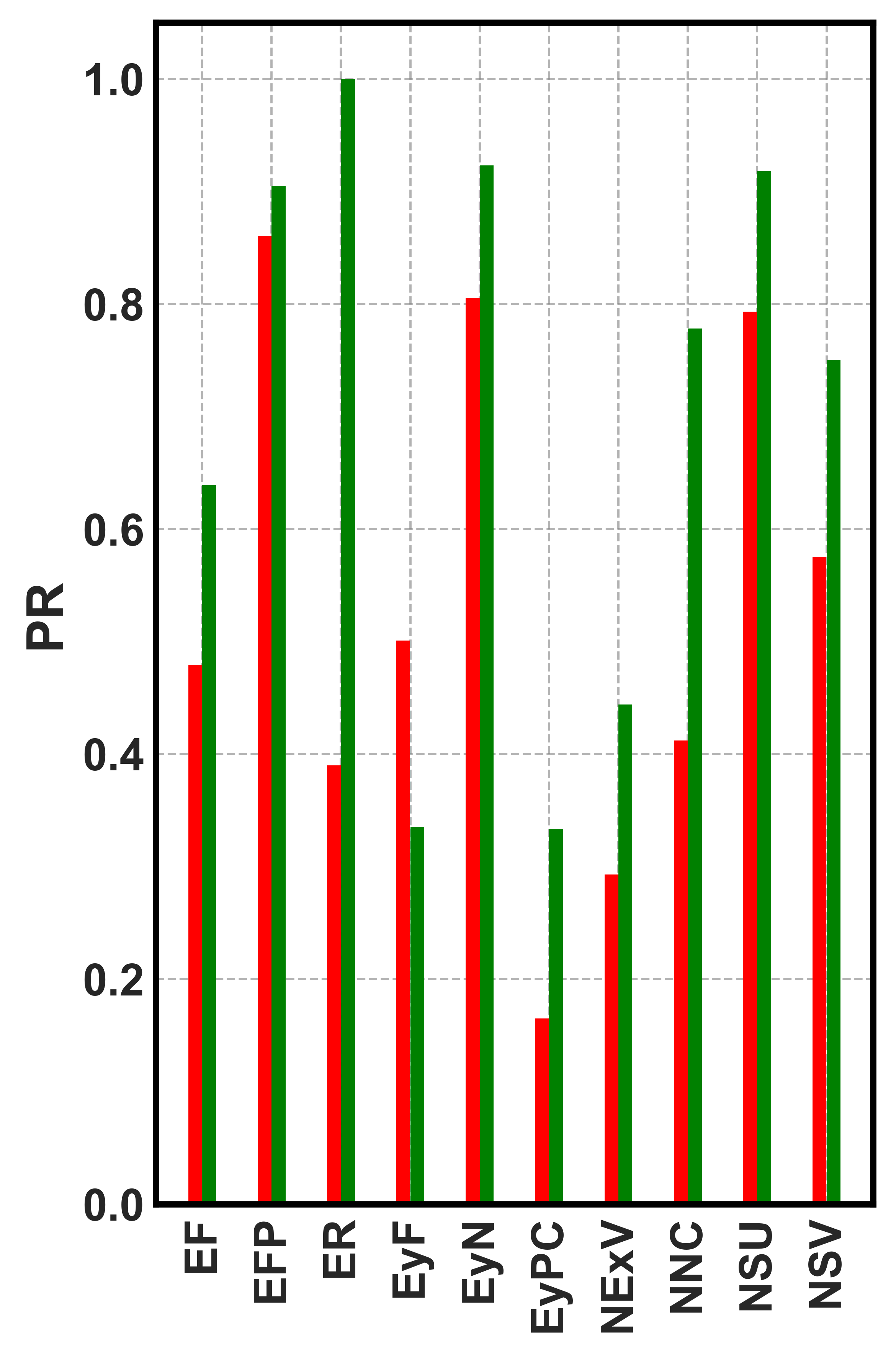}}
\subfloat[]{\includegraphics[width=1.7in]{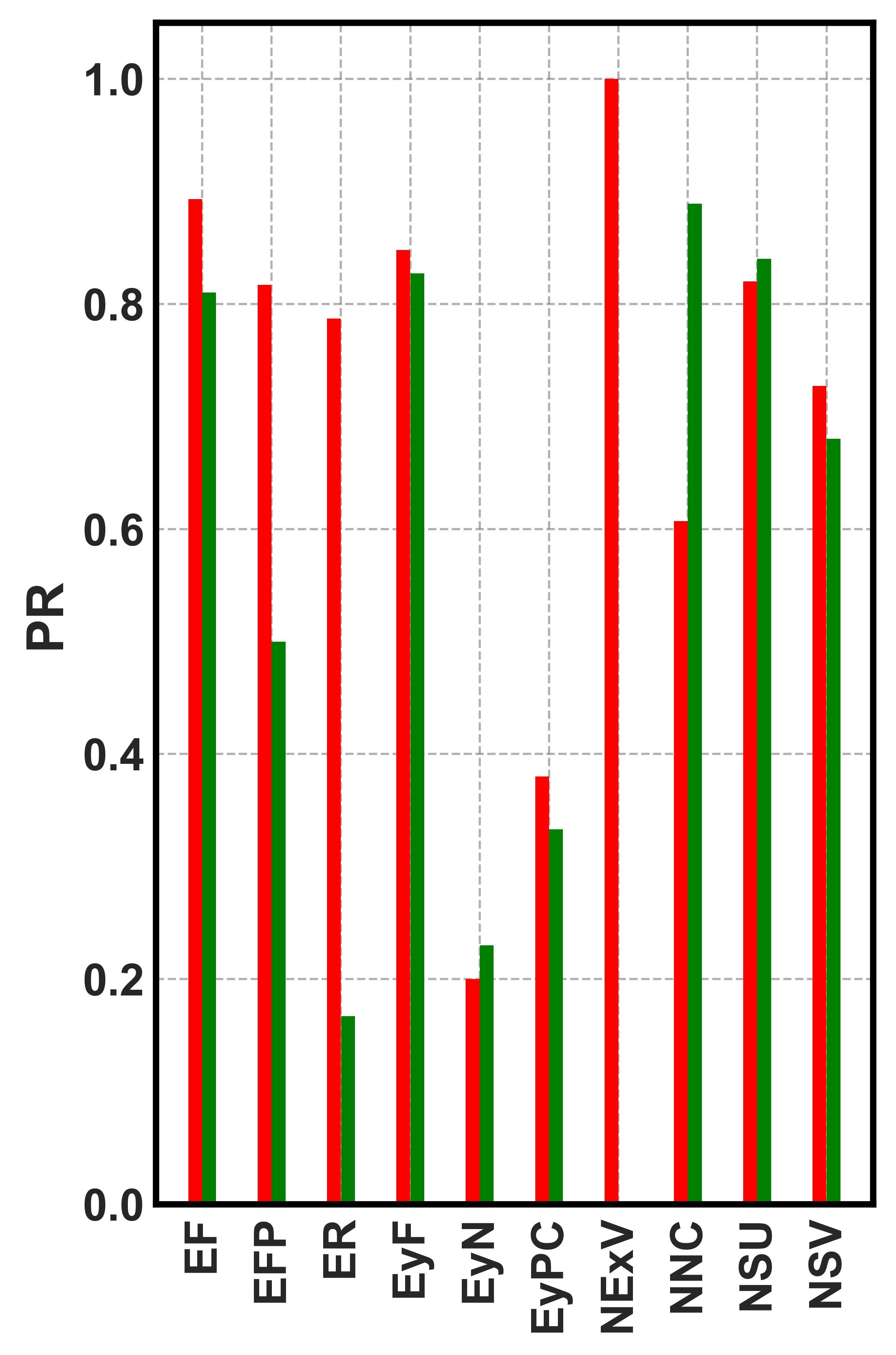}}
\subfloat[]{\includegraphics[width=1.7in]{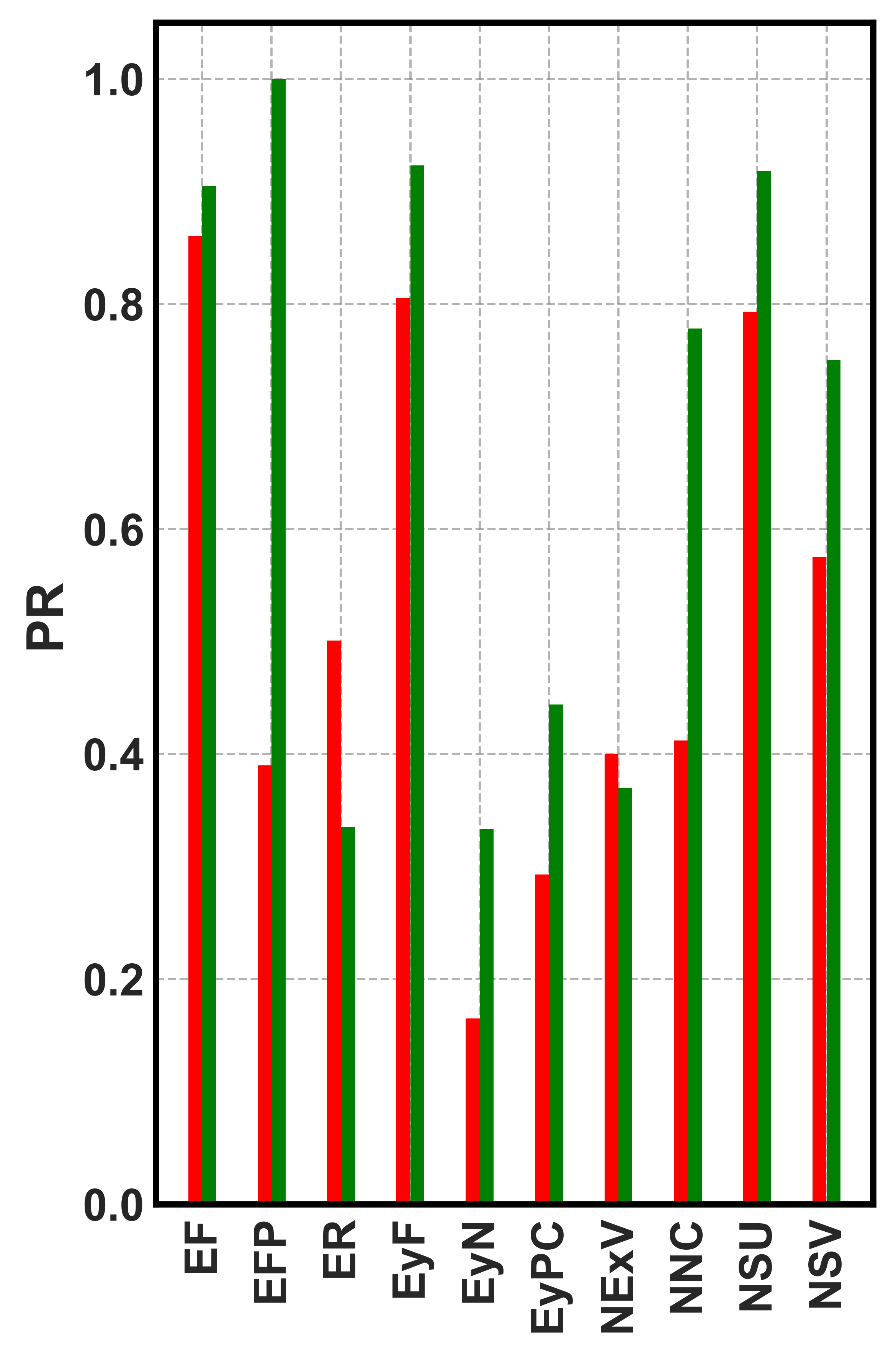}}
\subfloat[]{\includegraphics[width=1.7in]{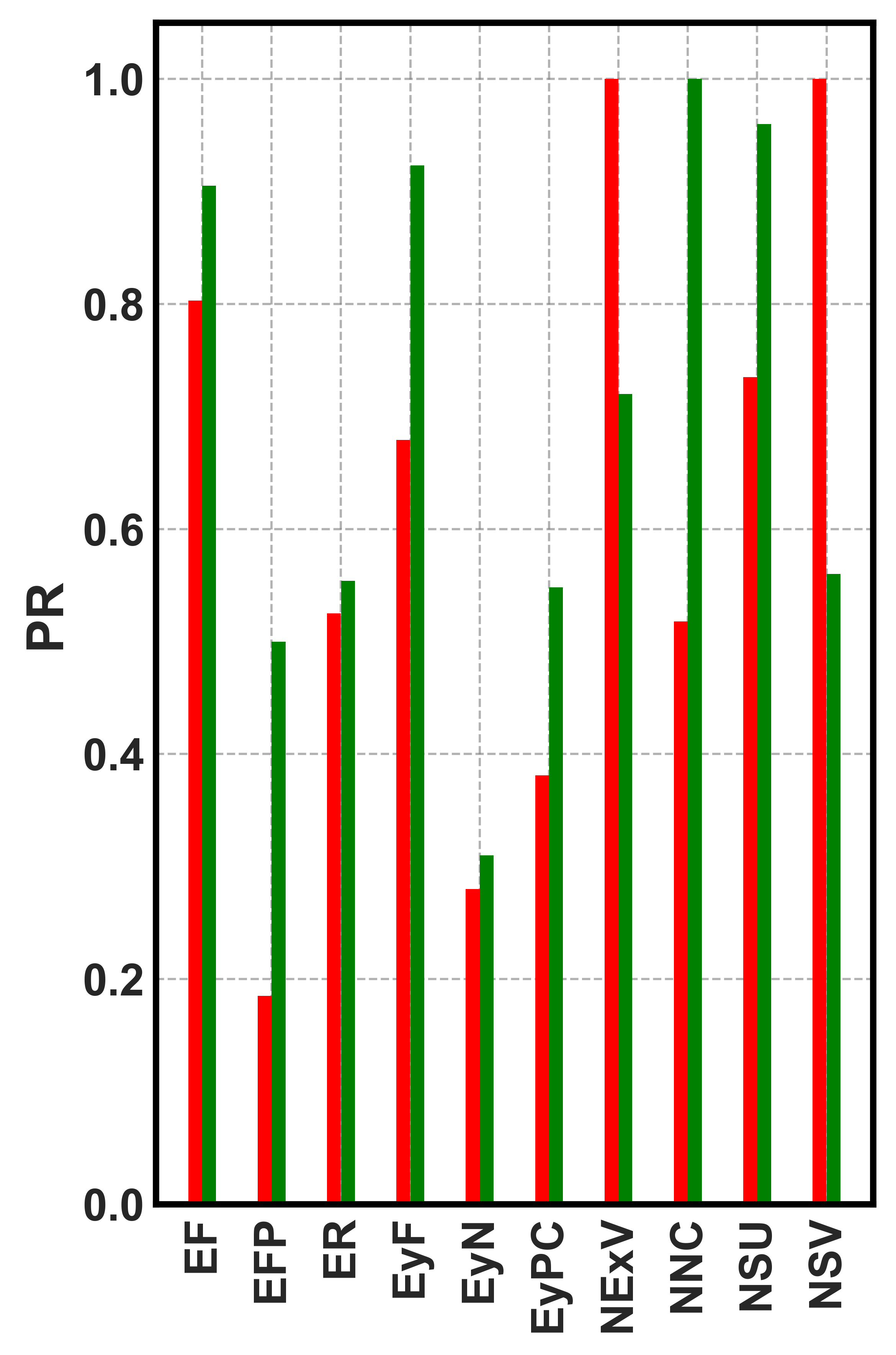}}
\caption{Classification performance (PR) of all detection models using precision (red) and recall (green) for the sheep facial landmarks: a) YOLOv5n, b) YOLOv6n, c) YOLOv7n, d) YOLOv8n, e) RetinaNet, f) EfficientDet, g) YOLO-Nas, and h) detectron2.}
\label{PR} 
\end{figure*}

In this section, we compare WGNN with different object detection models as shown in Fig. \ref{PR}. According to the sheep facial dataset, the precision and recall show that different facial landmarks perform differently. We also examine the effects of various detection models and the opportunities for clustering that arise from modifying these models. Furthermore, we assess the scalability and performance of our WGNN model on a mobile device.
\subsubsection{Baseline Facial Landmark Detection}
\begin{figure}[htbp!]
    \begin{center}
    \hspace*{-1em}
        \includegraphics[scale=0.59]{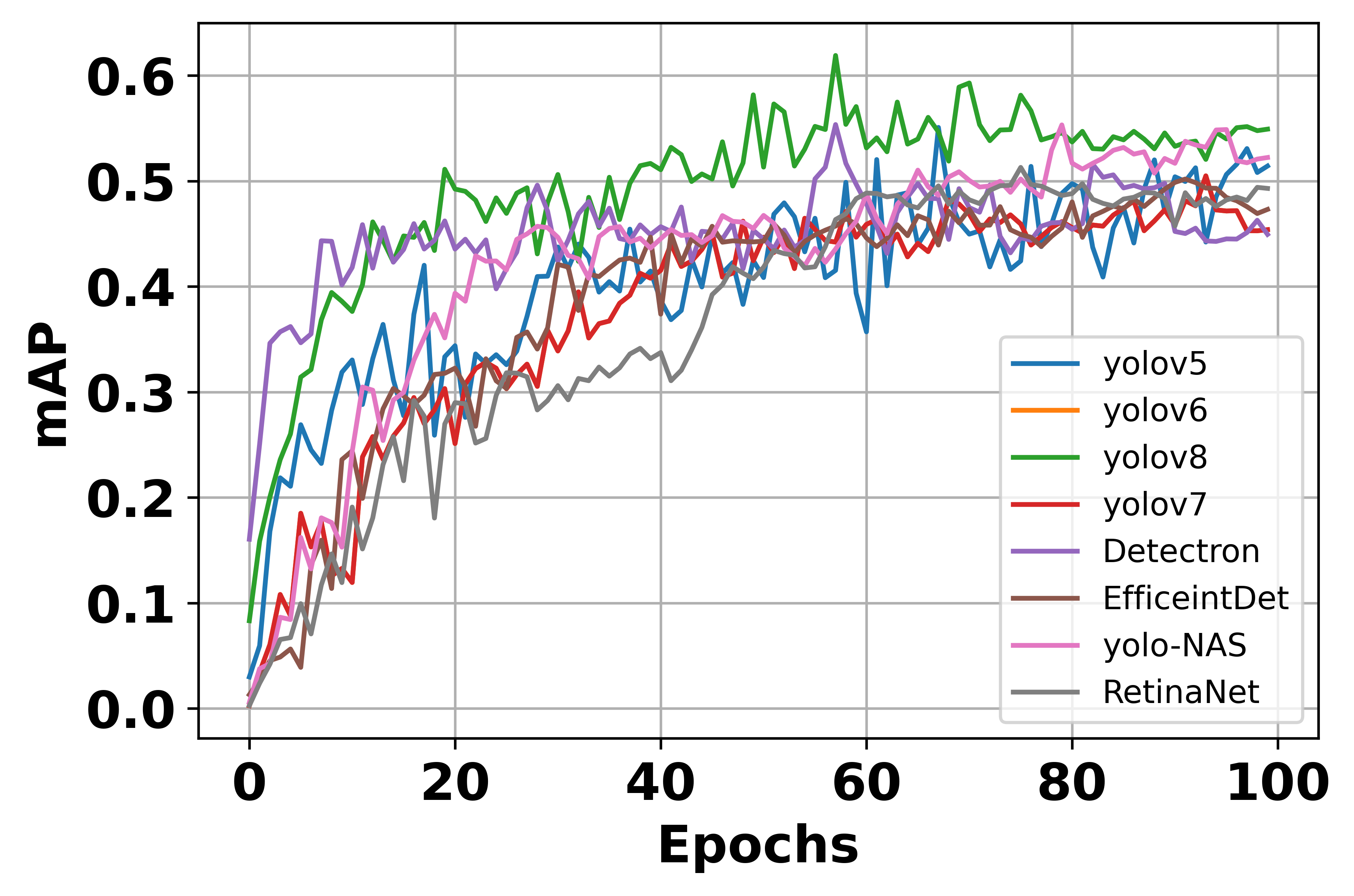}
        \caption{mAP for different facial landmark detection models and its evolution with training epochs. }
        \label{mAP_YOLOandALL}
    \end{center}
\end{figure}

Table \ref{yolo_comparison} displays the comparative performance of sheep facial landmark detection across the sheep facial landmark dataset using the top mAP-scored object detectors. We observe significant differences in detection performance across the different models, with the sheep facial landmark dataset achieving the best performance (59.3\% mAP) using YOLOv8n. With all images cropped in similar settings and only 9 types of facial landmarks present, this dataset boasts the most uniform set of image sequences. The mAP across the sheep facial landmark dataset remains consistent, with values ranging from 45.39\% to 59.3\% for five facial parts with different expressions using different object detection models as shown in Fig. \ref{mAP_YOLOandALL}. It is given that the detectors are learning to capture the landmark positions of sheep facial parts expressions rather than depending on the types of sheep information. However, the sheep facial landmark dataset produces a dramatic increase and decrease in precision and recall for different parts expressions of the face, as shown in Fig. \ref{PR}. The detection performance of the YOLO-NAS is comparable to that of the YOLOv5n (52.23\% mAP). Additionally, the low precision for different parts of facial expressions may be due to the low number of facial landmark instances in the training set (Fig.~\ref{dataset}). Ultimately, by utilizing  YOLOv8n and YOLO-NAS models, the facial landmark dataset attains a maximum mAP. This indicates that models YOLOv8n and YOLO-NAS are better at predicting tighter bounding boxes with a confidence score of 45. 
However, YOLOv8n has one order of magnitude less parameters (6.3M) than YOLO-NAS (90M), making the former the right choice for deployment onboard mobile devices. The inference time for the YOLOv8n is 1.9ms and 0.4ms postprocess per image. Moreover, other YOLO Nano models are even faster and lighter than YOLOv8n but achieve lower precision, as shown in Table~\ref{yolo_comparison}. Therefore, in environments where memory and computational power are limited (like in embedded systems), users may favor these lightweight models.

\begin{table}[htbp!]
\centering
\begin{tabular}{|c|c|c|c|c|c|}
\hline
\textbf{Model} & \textbf{Model Size} & \textbf{Par} & \textbf{P} & \textbf{R} & \textbf{mAP(\%)} \\ \hline
\textbf{YOLOv5n} &  1.93 MB &  1.8M  &  69.61 & 46.26 & 52.23 \\ \hline
\textbf{YOLOv6n} &  4.2 MB &  4.3M  &  45.30 &48.50 & 45.33\\ \hline
\textbf{YOLOv7n} &  4.45 MB &  4.4M  &  49.06 & 51.24 & 45.39\\ \hline
\textbf{YOLOv8n} &  6.3 MB &  6.2M  &  49.92 & 56.22& 59.30\\ \hline
\end{tabular}
\caption{Comparison of YOLO Nano Models with fewer parameters (Par).}
\label{yolo_comparison}
\end{table}

\subsubsection{Facial Pain Assessment and Ablation Study}
\begin{figure}[htbp!]
    \begin{center}
    \hspace*{-1em}
        \includegraphics[scale=0.59]{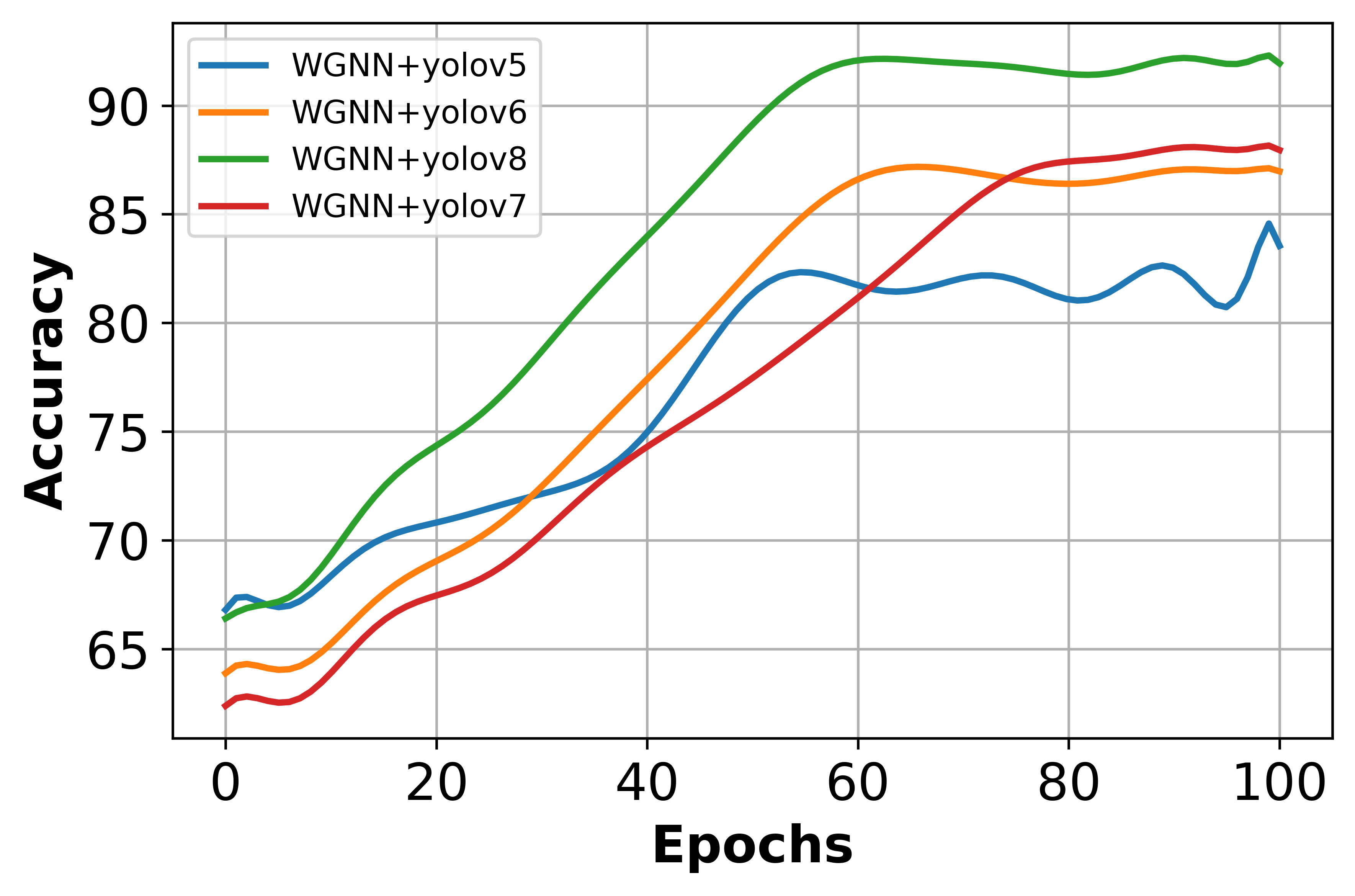}
        \caption{Accuracy after combining WGNN with the four different highest mAP-scored light-weight YOLO Nano models and its evolution with training epochs.}
        \label{yoloGNN_acc}
    \end{center}
\end{figure}

Fig. \ref{yoloGNN_acc} shows that the proposed WGNN framework training accuracy gradually increases as the number of training epochs increases. This is due to the WGNN module's utilization of detection models, which extract the facial landmark pain correlation from the sheep's face and estimate the hidden clustering states. YOLOv8n allowed achieving the highest training accuracy, namely 92.71\%. Even with unseen testing data, the model generalizes well and achieved 91.96\% testing accuracy. It means that WGNN and YOLOv8n can improve the aggregated pain score, which is calculated using a weighted average of the pain levels. Therefore, as the number of epochs increases, the detection models provide more facial landmarks, adequately training the proposed WGNN model. This indicates that we can achieve high accuracy due to the increased availability of high-quality data on facial landmarks with annotations.

\begin{figure*}[htbp!]
    \begin{center}
    \hspace*{0em}
        \includegraphics[scale=0.21]{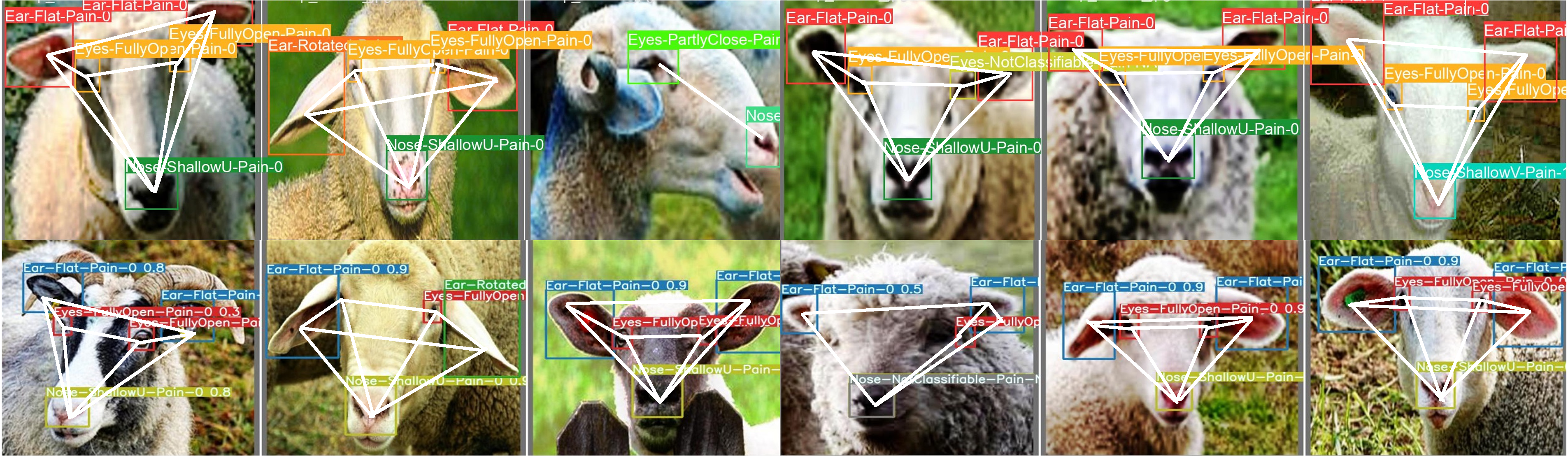}
        \caption{Evaluation results with WGNN applied after the object detection model, using the facial landmarks to find connections between these parts expressions using clustering.}
        \label{evaluations_results}
    \end{center}
\end{figure*}

For a fair comparison, 
all detection models use the same training parameters and batch size.
Fig. \ref{yoloGNN_acc} shows the ablation study of combining WGNN with all object detection algorithms with respect to the object detection accuracy for sheep facial landmark datasets. We create four pairs made of WGNN after an object detection model, namely YOLOv5n, YOLOv6n, YOLOv7n, and YOLOv8n. 
Specifically, we chose these YOLO models for their high mAP score and lightweight capability needed for mobile devices. 
WGNN can combine detected facial landmark expressions to look at the overall expression on the face and make performance clustering better. We also note that the YOLO approach mAP outperforms the scenario where WGNN models execute on each facial landmark, resulting in significantly enhanced clusters that accurately determine the pain level. In particular, the WGNN pairs with YOLOv8n and YOLOv7n are very complementary and achieve the best performance. Fig.~\ref{evaluations_results} shows some of the correct predictions with different colors and pain levels of WGNN combined with YOLOv8n for pain level assessment of sheep faces after training. These cases were not used in training and show how well the model generalizes and mitigates the bias of the same distribution.  

\subsubsection{State-of-the-art (SOTA) Comparison}
We compared the WGNN performance with three different SOTA models, as shown in Table~\ref{SOTA}: Support Vector Machine (SVM) \cite{7961768}, EfficeintNetB5 \cite{HIMEL2024200093}, combining convolutional and vision transformers (CCVT) \cite{LI2023107651} and convolutional neural networks (CNN) \cite{10.1145/3650400.3650652}. The WGNN model outperforms previous models. For example, SVM training accuracy performance is 71.55\% with evaluation testing accuracy results of 62.08\%, which overfits. The SVM is computationally expensive, and it is highly challenging to select the optimal regularization parameter. EfficientNetB5 achieved a training accuracy of 89.33\%. This model also uses multi-scale feature extraction layers or hierarchical convolutional layers to get hierarchical spatial global features from input grids of pixels. However, it encounters difficulties in assessing discrete feature relationships and the pain level of sheep faces. WGNN, on the other hand, exhibits greater robustness and accuracy when dealing with discrete landmarks since WGNN utilizes nodes and edge relationships to understand the interdependencies of facial landmarks, thus accurately assessing the pain level of sheep faces. On the other hand, the CNN model achieved just 79.15\% training accuracy, which might be due to weak performance in creating explicit contextual relationships between different sheep facial landmarks. The CCVT model achieved 85.45\% training accuracy along with 83.33\% testing accuracy. CCVT also needs grids of pixels for positional encodings of heatmaps to create spatial relationships of facial landmarks, which increases computational overhead.  

\begin{table}[htbp!]
\centering
\begin{tabular}{|c|c|c|}
\hline
\textbf{Model} & \textbf{Training Accuracy} & \textbf{Testing Accuracy}  \\ \hline
\textbf{SVM \cite{7961768} (2017)} &  71.55\% & 62.08\%   \\ \hline
\textbf{EfficeintNet \cite{HIMEL2024200093} (2024)} &  89.33\% & 86.60\%   \\ \hline
\textbf{CCVT \cite{LI2023107651} (2023)}& 85.45\% &  83.33\%   \\ \hline
\textbf{CNN \cite{10.1145/3650400.3650652} (2023)} &  79.15\% &  78.56\%  \\ \hline
\textbf{WGNN} &  \textbf{92.71\%} &  \textbf{91.96\%}  \\ \hline
\end{tabular}
\caption{WGNN comparison with SOTA models.}
\label{SOTA}
\end{table}
\subsubsection{Performance of Total Pain Score Comparison}
In the previous section, we assessed how WGNN employs redundant detections in conjunction with object detection models to enhance accuracy. In this experiment, we test the proposed WGNN model with YOLOv8n object detection models to find the total pain score (TPS) as given in \cite{MCLENNAN201619}. The WGNN reports results for sheep facial landmark datasets with five days time monitoring.
\begin{figure}[htbp!]
    \begin{center}
    \hspace*{0em}
        \includegraphics[scale=0.28]{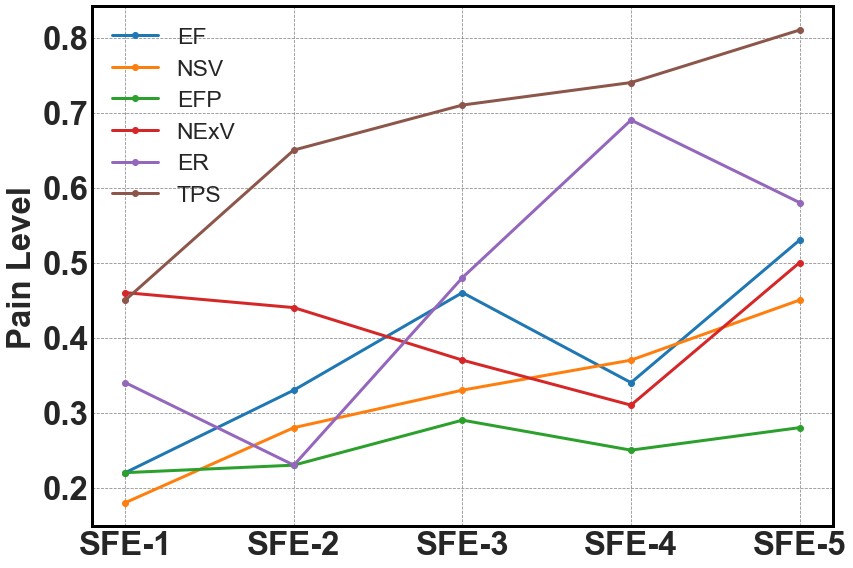}
        \caption{Correlation between the pain level 
        and the number of sheep facial expressions (SFE) used to track the total pain score (TPS) in time. The pain levels go from 0.1 (very low) to 1 (very high).}
        \label{painlevel_Days}
    \end{center}
\end{figure}

 Fig. \ref{painlevel_Days} shows the TPS for sheep facial expression (SFE) 
 evolving over the 5 days monitoring period. 
 Long-term assistance confirms that sheep experience pain due to a specific illness and do not rely on environmental factors. 

\subsubsection{Limitations of Study}
\begin{figure}[htbp!]
    \begin{center}
    \hspace*{0em}
        \includegraphics[scale=0.135]{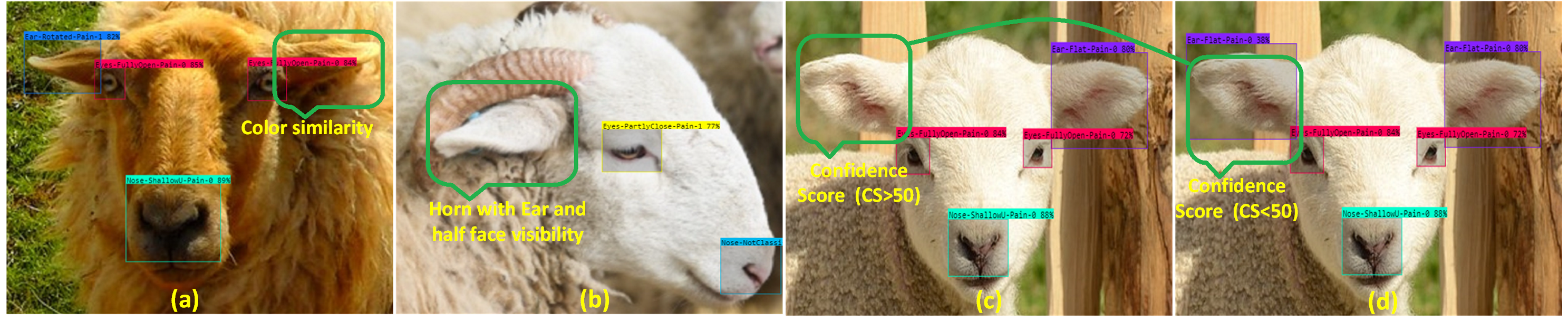}
        \caption{Cases of incorrect sheep facial landmark detections due to color similarity, half-face visibility, ear under horn, and low confidence score.}
        \label{Limitations}
    \end{center}
\end{figure}

The performance of the detection algorithm determines the proposed WGNN model. Fig. \ref{Limitations} shows examples of situations that led to false pain predictions. These include higher color similarity, which affects the object detection performance later, affecting the proposed WGNN model performance; half-face visibility during face detection; this factor can be mitigated using mobile devices such as UAVs; and a low confidence score since obtaining the desired TPS accuracy may require more than 50 confidence scores in landmark detections. These factors will be taken into account in future work.

Moreover, we also plan to collect more data from different parts of the world with different characteristics and varied environments to ensure generalization, with different distributions across the world and different sheep types with colors and sizes. The increase in the number of data collections will not only mitigate the overfitting problem of generalization with different distributions but could also improve the performance of the WGNN training and testing accuracy. Moreover, in the proposed study we focused on the adult sheep, which might create a low confidence score for the lambs as shown in Fig. \ref{Limitations}. SPFES parameters are defined for the adult sheep, but lambs parameters of facial expression should be considered in future studies for more advanced-level dataset training with WGNN. 
\section{Conclusion}\label{Conclusion}
This article proposed coupling a WGNN model with an object detector model designed to detect clusterings and assign a TPS to the facial expressions of sheep. The design caters for scenarios that pose challenges in capturing real-time facial expressions in their natural states. We design the detector model to identify facial landmarks and pinpoint the location of each facial part expression, utilizing the proposed sheep facial landmarks dataset. Next, we proposed using a WGNN to enhance the significance of each part feature expression, establish clusters based on their correlations, and compute an aggregated pain score by averaging the pain levels of the part expressions within each cluster. The detector model detection results emphasize the WGNN model performance. We show that using the sheep facial landmark dataset can improve the proposed model's overall automated performance in figuring out how much pain a sheep is in. We conducted studies using authentic sheep facial expression images and a limited quantity of training data. There is evidence that our proposed sheep facial landmarks dataset is better in terms of mAP using detector models and accuracy to find the TPS with the proposed WGNN model, especially when the training data available is limited.
\section{ACKNOWLEDGMENTS}
This work was supported by the CISTER Research Unit \texttt{UIDP/UIDB/04234/2020} and project ADANET \texttt{PTDC/EEICOM/3362/2021}, financed by National Funds through FCT (Portuguese Foundation for Science and Technology). Also, this article is a result of the project supported by Norte Portugal Regional Operational Programme (NORTE 2020), and LASI \texttt{LA/P/0104/2020}.

\section{Ethical Impact Statement}
The proposed research work study is on improving animal welfare in the precision agricultural domain by developing a novel approach to monitor sheep health through facial expression analysis. Our work is expressly to assist farmers and veterinarians in early diagnosis and timely intervention in sheep health, potentially reducing the spread of illnesses within herds. Using mobile devices such as UAVs to identify sheep facial images, our system could promote more accessible and cost-effective animal health monitoring, especially in rural or remote areas with limited access to veterinary services. We acknowledge several ethical considerations associated with this research. First, in order to minimize any disturbance to the sheep's natural behaviors, the use of UAVs for sheep monitoring is subject to animal welfare rules. It is crucial that the model we propose remain non-invasive; we established a high value on the natural, stress-free analysis of animals in order to prevent the introduction of additional stresses that could affect the sheep as well as the farm ecosystem overall.

Moreover, data security and privacy are carefully considered when gathering and analyzing data from sheep, especially when using technologies like UAVs to collect visual data. Our approaches are made to protect the processing and storage of data and ensure the information collected is only utilized to advance the health and welfare of animals. To prevent false positives and negatives that might impact farm decision-making, we focused on monitoring the degree of pain experienced by animals top priority when proposing our models. By carefully and morally assessing the model and taking into account the possible consequences of incorrect diagnoses, we reduce the likelihood that misinterpretation of the sheep's health condition will result in inappropriate interventions.


{\small
\bibliographystyle{ieee}
\bibliography{egbib}
}

\end{document}